\documentclass[sn-basic]{sn-jnl}

\usepackage{natbib}

\usepackage{graphicx}
\usepackage{color}
\usepackage{lipsum}
\usepackage{colortbl}
\usepackage{wasysym}
\usepackage{tabularx}
\usepackage{listings}
\usepackage{subcaption}
\usepackage{multirow}
\usepackage[
  separate-uncertainty = true,
  multi-part-units = repeat
]{siunitx}
\usepackage{booktabs}

\usepackage{comment}

\newcommand{\redacted}{\texttt{[redacted for anonymity]}}
\newcommand{\anonymity}[1]{\ifthenelse{\equal{\version}{review}}{\redacted}{#1}}

\def\lacourpreview{\burl{https://www.trusthlt.org/lacour/}}
\def\lacourhuggingface{\burl{https://huggingface.co/datasets/TrustHLT/LaCour}}
\def\lacourgithubcorpus{\burl{https://github.com/trusthlt/lacour-corpus}}
\def\lacourgithubcreation{\burl{https://github.com/trusthlt/lacour-generation}}
\def\lacourqando{\burl{https://github.com/trusthlt/lacour-questions-opinions}}
\def\lacourhuggingfaceqando{\burl{https://huggingface.co/datasets/TrustHLT/QandO}}

\jyear{2023}%

\begin{document}
	
\title[LaCour!: Enabling Research on Argumentation in Hearings of the ECHR]{LaCour!: Enabling Research on Argumentation in Hearings of the European Court of Human Rights}

\author*[1]{\fnm{Lena} \sur{Held}}\email{lena.held@tu-darmstadt.de}
\author[2]{\fnm{Ivan} \sur{Habernal}}
\affil{\orgdiv{Trustworthy Human Language Technologies}}
\affil[1]{\orgdiv{Department of Computer Science}, \orgname{Technical University of Darmstadt}, \orgaddress{\country{Germany}}}
\affil[2]{\orgdiv{Research Center Trustworthy Data Science and Security of the University Alliance Ruhr}, \orgname{Faculty of Computer Science, Ruhr University Bochum}, \orgaddress{\country{Germany}}}

\abstract{
Why does an argument end up in the final court decision? Was it deliberated or questioned during the oral hearings? Was there something in the hearings that triggered a particular judge to write a dissenting opinion? Despite the availability of the final judgments of the European Court of Human Rights (ECHR), none of these legal research questions can currently be answered as the ECHR's multilingual oral hearings are not transcribed, structured, or speaker-attributed. We address this fundamental gap by presenting LaCour!, the first corpus of textual oral arguments of the ECHR, consisting of 154 full hearings (2.1 million tokens from over 267 hours of video footage) in English, French, and other court languages, each linked to the corresponding final judgment documents. In addition to the transcribed and partially manually corrected text from the video, we provide sentence-level timestamps and manually annotated role and language labels. We also showcase LaCour! in a set of experiments that explore the interplay between questions and dissenting opinions. Apart from the use cases in legal NLP, we hope that law students or other interested parties will also use LaCour! as a learning resource, as it is freely available in various formats at \lacourhuggingface.
}


\maketitle

\section{Introduction}

What can we learn about law and legal argumentation from court judgments alone? Contemporary research addresses empirical legal questions (e.g., which arguments are used) or legal NLP questions (e.g., predicting case outcomes) by relying on the availability of the final `products' of each case, the court decisions \citep{habernalMiningLegalArguments2023,medvedevaUsingMachineLearning2020}. The European Court of Human Rights (ECHR) is a prominent data source, as its decisions are freely available in a large amount, along with the metadata of the violated articles and other attributes. This makes ECHR a popular choice among NLP researchers \citep{aletrasPredictingJudicialDecisions2016,chalkidisLEGALBERTMuppetsStraight2020}. However, whether or not the legal arguments in ECHR's cases are created as a part of legal deliberation or are created post-hoc after reaching a decision remains an open (and partly controversial) question.

In order to better understand the legal argument mechanics, that is which arguments of the parties were presented, discussed, or questioned, and thus might have influenced the case outcome, we must take the oral hearings into account. We witness that the availability of oral hearing transcripts of the U.S. Supreme Court enables further legal research \citep{ashleyLearningDiagrammingSupreme2007, fang-etal-2023-super,fang-etal-2023-votes}. However, empirical research into the interplay of arguments at the court hearings and the final judgments has been so far impossible for the ECHR, as there are no hearing transcripts available.

This paper addresses this research gap. Given only the raw video material recorded during selected ECHR hearings, we ask to which extent can we semi-automatically create a high-quality multi-lingual corpus that would enable further empirical legal NLP research. The challenges that we need to tackle include second-language speakers in legal jargon, missing speaker attribution (which judge or party presents their arguments/questions and when), low-quality court videos with technical glitches, and no alignment between the hearings and the final published court decisions.
We address these challenges by combining automatic speech recognition with extensive manual corrections and annotations and create transcripts that are structured, as accurate as possible and, last but not least, easy to read.

We present LaCour!, the first corpus of partly manually-corrected and speaker-aligned multi-lingual textual transcripts of hearings of the ECHR. To showcase an exemplary scenario in which this corpus can be used, we study the correlation between the interactive part of a hearing and the formation of a judicial opinion in the judgment. More specifically, we focus on the connections between the questions posed by the judges to the parties during the hearings and the possible judicial opinions of these judges later on. We would like to know explicitly whether it is possible to see a foreshadowing of this opinion already during the hearing.

We hope that LaCour! will open up new research possibilities for the NLP community to look at the whole process of a case. In addition the transcripts can also benefit the empirical legal community, which may wish to study specific hearings in written form along with the corresponding judgments.

\section{Related work}

As legal NLP is a large subfield of NLP covering a wide range of topics, here we focus mainly on works dealing with hearing transcripts. We cover both the U.S. Supreme Court as well as the ECHR.

\subsection{Hearings of the U.S. Supreme Court}

For the U.S. Supreme Court there are two major resources providing the public with material on oral hearings. The first one is the \emph{Oyez}\footnote{\burl{https://www.oyez.org/}} website, which is the brainchild of Jerry Goldman~\citep{goldmanPoliticalScienceMultimedia1998}. The website contains and curates oral hearings before the U.S. Supreme Court as audio and textual versions and includes entries that date back to 1955. Since its original creation, the project has continued as a collaboration between Cornell's Legal Information Institute, Justia and Chicago-Kent College of Law. The Oyez Team has taken great effort to transfer material from the pre-digitial era into a digital and publicly available format. Most of the source material has been curated and enriched with additional information, like audio-aligned transcripts, case summaries and Supreme Court opinions.
\citet[p.~34]{goldmanPoliticalScienceMultimedia1998} highlights that it is crucial to have the possibility to actually listen to a hearing itself, because it conveys more meaning and nuances than only a transcript would, as these recordings ``are rich in emotive content, which may have a material bearing on the attitudes and directions in a justice's line of argument." The task of transcription was delegated to professional companies.\footnote{\burl{https://www.oyez.org/about} cites three companies, Heritage Reporting, Tech Synergy and Alderson Reporting for contributing transcriptions}
All recordings from 1955 onwards have been digitized, resulting in a total of more than 14,000 hours of audio material and over 66 million words.\footnote{\burl{https://www.oyez.org/history}} The database was intended as a learning and researching resource for students as well as legal scholars.

The second important resource for material concerning the U.S. Supreme Court hearings is \emph{courtlistener}\footnote{\burl{https://www.courtlistener.com/}} which is part of the Free Law Project.\footnote{\burl{https://free.law/about}} The site provides access to legal opinions, various documents and oral argument recordings from federal, state and specialty courts. Although oral arguments are available in digital format, no transcripts are included. There are also other websites curating legal data for the public, like the Supreme Court Database~\citep{Spaeth2022} or the Caselaw Access Project,\footnote{\burl{https://case.law/}} however, they do not include oral arguments in their provided material.

Easily accessible and extensive collections of all court material naturally play an important role for legal scholars. But the transcripts alone are also of great importance for empirical research, as they enable such research for NLP in the first place. Some approaches have used the transcripts of the oral arguments to gain insights into the argument structure for the purpose of tutoring~\citep{ashleyLearningDiagrammingSupreme2007}, others have tried stance detection~\citep{bergamLegalPoliticalStance2022} or machine learning-based approaches to study the judges’ attitudes~\citep{dickinsonComputationalAnalysisOral2018}. More recently~\citet{fang-etal-2023-super} compiled a large collection of court documents of the U.S. Supreme Court, called Super-SCOTUS, which specifically highlights the role of hearing transcripts. In the work by~\citet{fang-etal-2023-votes}, Super-SCOTUS enables the study of the role and influence of partisanship in oral hearings. 

\subsection{Hearings of the ECHR}
On the European side of legal analysis in the context of NLP, a popular choice of investigation is the ECHR.\footnote{\burl{https://www.echr.coe.int/}} This court is a frequently used source for analysis, not only because of the availability of its documents, but also because of its importance and reputation as a supranational body. It has been in existence since 1953 and deals with applications brought by individuals or groups, in rare cases even states, against states that are contractors to the European Convention on Human Rights (``The Convention"). The Convention contains several articles\footnote{\burl{https://www.echr.coe.int/european-convention-on-human-rights}} concerning different aspects of human rights and applications can refer to multiple of those articles. The court receives over 50,000 applications every year which are processed according to importance and urgency. For roughly 30 Chamber or Grand Chamber cases each year, the court holds a public hearing which is recorded.\footnote{more information on how an application is processed can be found in \burl{https://www.echr.coe.int/documents/d/echr/your_application_eng}}

In contrast to the good availability of the U.S. Supreme Court argument transcripts, no such data is available for the ECHR. The ECHR solely publishes judgments or decision through its HUDOC database.\footnote{\burl{https://hudoc.echr.coe.int/}} For this reason, most works pose research questions that can leverage information available in the HUDOC data collection.

A frequently pursued line of research examines the argumentative nature of the judgments \citep{mochales-palauStudySentenceRelations2007,mochalesCreatingArgumentationCorpus2009,mochalesArgumentationMining2011a,poudyalECHRLegalCorpus2020}; we refer to a recent work by \citet{habernalMiningLegalArguments2023} for a survey of this research area.

Another popular use of ECHR material is legal judgment prediction, most notably investigated by~\citet{aletrasPredictingJudicialDecisions2016}. Besides classifying cases in a binary manner for violation of a specific Article, they also introduce a publicly available corpus with 584 cases. All cases in this dataset are in English, relate to Articles 3, 6 and 8 of the Convention and are balanced in regards to violation and non-violation. \citet{chalkidisNeuralLegalJudgment2019} collect 11.5k cases in a similar manner and use them for binary classification, multi-label classification and case importance detection with the help of neural models. \citet{medvedevaUsingMachineLearning2020} use a similarly sized portion of ECHR cases and experiment with only using information available before the final judgment to simulate actual prediction rather than classification. In an additional experiment, they also investigate a connection of a specific judge being part of the Chamber and the decision of the case. The latter experiments even achieved a surprisingly high correctness for certain Articles of the Convention. However, prediction for ECHR cases still suffers from the issue that argumentative text is only available after the judgment has been passed and anything using the judgment documents is already biased towards the outcome. Moreover, as \citet{villataThirtyYearsArtificial2022a} pointed out in their review of \citep{medvedevaUsingMachineLearning2020}, classifying legal cases is a different and more complex matter than a simple binary classification.

Finally, ECHR documents are also a popular source used in large dataset collections or benchmarks \citep{chalkidisLexGLUEBenchmarkDataset2022, hendersonPileLawLearning2022,guha-etal-2023-legalbench}, which made them a standard and go-to option for many works in legal NLP. And due to the amount of available text, they have also been used as training material for domain specific language models, like LEGAL-BERT~\citep{chalkidisLEGALBERTMuppetsStraight2020} which have opened up new possibilities for handling legal texts. 

A case in the ECHR can be evaluated on more than the final decision. Factors outside of the content of a decision, such as voting behavior, opinions and circumstances which potentially influence a judge's decision have also been studied empirically using decisions and opinions taken from HUDOC~\citep{voetenImpartialityInternationalJudges2008}.

Still, NLP-based techniques such as legal argument mining and legal judgment prediction focus on the written documents provided by the ECHR through HUDOC. Because there are no transcripts of the public oral arguments in hearings in the ECHR available, to the best of our knowledge, no such research has been conducted in this area and although the hearings are publicly available, they have not been scrutinized in NLP research to date. The hearings also present a data gap in the current representation of legal proceedings in the ECHR. Although held only on a proportionally small number of cases, oral hearings can provide insight into the process that takes place between an application and the final judgment.
\begin{figure}
	\centering
    \includegraphics[width=.4\textwidth]{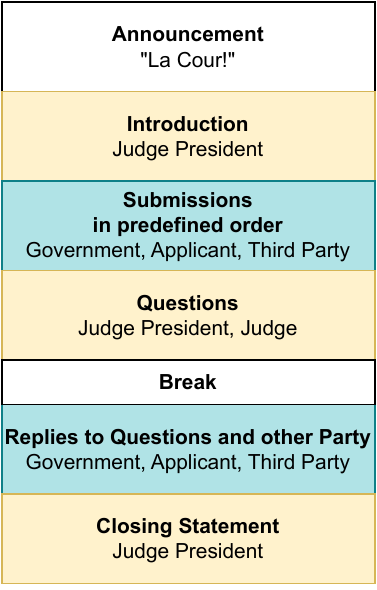}\hfill
	\caption{An illustration of a typical hearing structure. Variations in the structure are possible, e.g. occasionally the announcement is missing or no questions are asked}
	\label{fig:hearingstructure}
\end{figure}
\section{Creating the LaCour! corpus}
A typical hearing before the European Court of Human Rights is structured into several phases which are outlined in Figure~\ref{fig:hearingstructure}. The judge president opens the hearing, introduces basic information concerning the case and acts as a moderator throughout the session. They also announce the predetermined order of addresses and call the first party. Their job is also to remind the speakers when their allocated time has passed. After each party has taken their turn to address the Court and present their interpretation, the judges are called to ask questions to the parties. In earlier hearings, the judge president would then ask the parties to immediately reply to those questions as well as to the other party’s remarks, but in more recent hearings, a short break is usually permitted to allow for some preparation. In the most recent hearings there is even an opportunity for the judges to ask follow-up questions again after the replies to the first round of questions. After these replies, the judge president delivers the closing statement. This debate-like, moderated structure gives the hearings a ceremonial character. But nevertheless, the language used is different, as it is not as formal as that of written legal texts, even though “Legalese” is still present in the vocabulary used.

\subsection{Hearing videos}

As our source material, we collected all videos of hearings - called webcasts - that are available on the website.\footnote{\burl{https://www.echr.coe.int/webcasts-of-hearings}} At the time we started the collection, the available hearings ranged from 2007 to 2022 with 3 different audio tracks: the original version without interpretation, an English interpretation and a French interpretation. There are no official transcripts of the hearings and the videos do not contain any subtitles. The only available material are live recorded videos with the live recorded interpretations as an optional audio track.

A single hearing is not necessarily held in a single language. Besides the official languages of the Court, English and French, participants are often allowed to speak in their native language. Consequently, a certain percentage of hearings (around 20\% in our subset) are not held in one or two, but several spoken languages. Both the English and French interpretation audio tracks are recordings of the live interpretation which were created at the time of the hearing and do not sound as fluent as the original. Although we have no way of knowing which version a participant or observer heard at the live hearing, we chose to focus on the original version, which contains no interpretation or translation. Our initial collection contained 289 hearings between 2007 and 2022 with the original audio in various languages at the time we scraped the website.

We probed the quality of the videos and noticed that most videos before 2012 are perceived as worse than recent ones and prone to technical errors, like sound dropouts, microphone failures and incorrect audio tracks. A preliminary transcription test with some of the hearings of the early years also revealed a poor quality. In addition, there are several hearings with parts of the hearing missing or cut off. Therefore, we decided to only use hearings starting from the year 2012.

\begin{figure}
	\centering
    \includegraphics[width=1\textwidth]{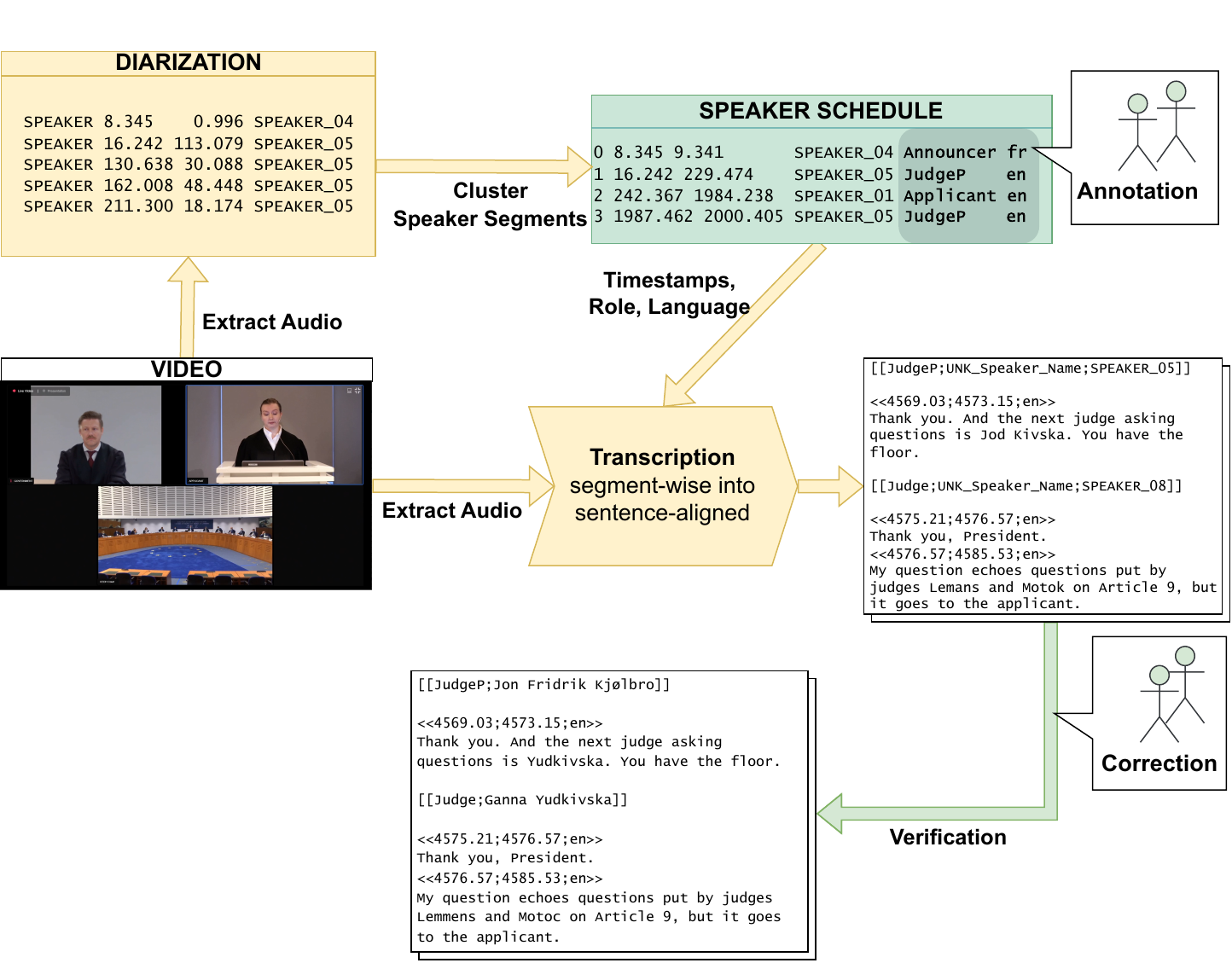}\hfill
	\caption{Illustration of the transcript generation pipeline. Yellow elements indicate an automatic processing, while green elements indicate manual annotation and correction. The associated code is available on \lacourgithubcreation}
	\label{fig:pipeline}
\end{figure}

\subsection{Generating transcripts}
Getting from a pure video to a structured and annotated transcript required the design and implementation of a processing pipeline. The very first step in said pipeline, illustrated in Figure~\ref{fig:pipeline}, is the speaker diarization, which is applied to obtain a schedule of who is speaking at which time. The diarization also takes care of overlapping speech that occurs in rare cases. For this task, we ran \texttt{pyannote}\footnote{\burl{https://huggingface.co/pyannote/speaker-diarization}}~\citep{Bredin2020} which returns a fine-grained schedule of speech segments and a supposed speaker identification number. As a next step, we clustered the output by a change in speaker into larger chunks to create a ``speaker schedule” that resembles the order of speeches in the hearing.

A particularly hard step in the pipeline was the language detection. We originally wanted to use automatic language detection to tag each segment of the speaker schedule. However, initial trials on selected samples with several \texttt{wav2vec2}~\citep{baevski-2020-wav2vec2-neurips} models\footnote{We experimented with the huggingface models \texttt{facebook/wav2vec2-conformer-rel-pos-large-960h-ft}, \texttt{facebook/wav2vec2-base-10-voxpopuli-ft-en}, \texttt{facebook/hubert-large-ls960-ft}} to detect the spoken language did not yield sufficiently reliable results. Even the much more powerful Whisper~\citep{radford2022whisper} with an in-built language detection in the pipeline had issues with many instances that we tested. One major problem is the fact that participants of the hearings, especially the judges, tend to speak both English and French and not seldom use both languages in a single sentence. A common example of this is “Merci beaucoup, Madame. Judge [Name], please.” The second most common problem is related to the language proficiency of the speaker. With only English and French being the official languages, many speakers before the Court do not speak in their native language. For a human, this is not a challenging task, automatic language identification models on the other hand had serious problems and frequently picked up the native language of the speaker, rather than the actual spoken language. As an example, French speeches by Italian native speakers were classified as Italian. We suspect that the detection models are biased towards native speakers, which could be caused by the data they are trained on, which is mostly clean speeches spoken by native speakers.

We could not find a reasonable way to automatically label the language, without having to check every instance. We therefore decided to \textbf{manually identify} the languages for each section in our coarse-grained speaker schedule. To achieve this, we hired two research assistants who were trained to listen to parts of the segmented videos in order to identify the spoken language. They validated the correctness of the timestamps and adjusted them if necessary by directly modifying the schedule file. Simultaneously, they tagged the affiliation of the speaker, which would later be needed to distinguish a party's address from the judges' questions. The correct role could be inferred from nameplates in the video and the context of the president's announced order. Any conflicting identification by the annotators was discussed until a correct tag was agreed on. This manual step also allowed us to correct small diarization errors. The resulting speaker schedules contain segments which are marked with timestamps, as well as the spoken language in ISO 639-1 format and the role of the speakers. For the roles, we decided to distinguish between the parties appearing in court. These are `Announcer', `Applicant', `Government', `JudgeP' (judge president), `Judge' and `Third Party'. In cases in which there are multiple parties, as for example multiple applicants, the tags do not distinguish between them. In the case ``Savickis and Others v. Latvia (no. 49270/11)" for example, the representatives for ``Savickis" and the representatives for all other applicants would be tagged as ``Applicant". 

This tagged structure is sufficient, but also necessary for the central component of the pipeline: the transcription. Our method for generating a transcript is based on Whisper\footnote{\url{https://huggingface.co/openai/whisper-large-v2}}~\citep{radford2022whisper} on a single entry of a speaker schedule, passing the previously identified language as a parameter. Passing the language tag circumvents the in-built language detection and ensures that that the output text is in the correct language. Inputting the typically long segments of speech results in a single large block of text. To ensure better alignment between timestamps and transcripts, we used \texttt{stable-ts}\footnote{\url{https://github.com/jianfch/stable-ts}} which generates more accurately aligned sentence segments of the transcript. As a result, the transcript is not just divided into speaker segments, but also into fine-grained sentence segments.
A last manual correction step was necessary to properly separate instances of speaker overlap, as the model generates inaccurate transcripts when two speakers talk at the same time. We identified segments with potential overlap automatically using the timestamps and the annotators were tasked to separate and resolve these conflicts. 

\begin{figure}
    \centering
    \includegraphics[width=.9\textwidth]{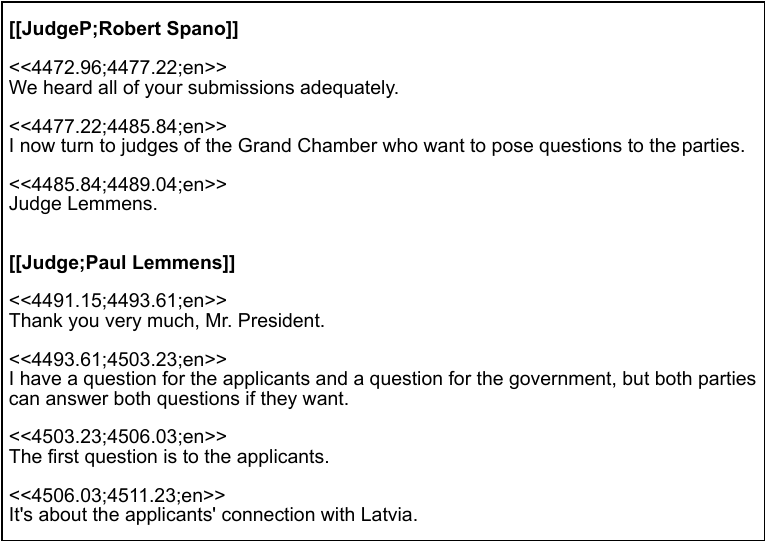}
    \caption{Excerpt of the transcript to webcast of Savickis and Others v. Latvia (no. 49270/11), webcast id \texttt{4927011\_26052021}}
    \label{fig:transcript_excerpt}
\end{figure}

Even though this transcript version is sufficient for understanding the hearings, we decided to further refine it by adding the names of the judges and correcting the transcript. To obtain the names, we gathered lists of hearing participants from linked documents in the HUDOC database titled ``Forthcoming Hearing" and verified that the listed participants also appear in the ``Judgement" document. Judges and names in general are often misspelled in the transcript and we used the listed participants to correct the spelling. This correction enabled us to later link judges' opinions in the final judgments with questions they asked during the hearing.

We hired two research assistants with a high proficiency in English and French to manually correct all parts spoken by the judges or the judge president in the automatically generated transcripts. The research assistants reported an overall good quality and little corrections to be made. Legal terminology and named entities were occasionally inaccurately transcribed. For example, domain specific terminology such as ``tax arrears" was transcribed as ``tax errors". Corrections took them approximately 1.5 hours per hearing on average, resulting in 300 hours invested for all hearings. Due to limited resources and issues in finding a suitable crowdsourcing opportunity, speeches by other speaker roles (``Applicant", ``Government", ``Third Party") and in other languages besides English and French remain uncorrected. In parallel, we instructed the research assistants to add the names of the judges to the corrected segments. They were able to chose from the previously compiled list of attendants to make sure that there are no differences in spelling.

Finally, our completed transcript corpus comprises tagged speech segments with tagged sentence-aligned transcriptions, which are manually corrected for the labels ``Judge" and ``Judge President", but uncorrected for all other labels. Figure~\ref{fig:transcript_excerpt} shows an excerpt of one transcript. Obtaining the final partially corrected transcript from the video version required the work of 4 student research assistants and took around 10 months. We call our new corpus \textbf{LaCour!} in honor of the traditional and catchy announcement at the beginning of each hearing.

\subsection{Linked documents}
In order to represent a large part of the entire process that surrounds a hearing and case as adequately as possible, the hearing needs to be considered in its context. We therefore also collected the most important documents that accompany a hearing as a supplementary text resource. Although documents relating to written communication prior to a hearing are scarce, there is almost always a concluding document containing the ``Judgement" by either the Grand Chamber or the Chamber. Since those documents summarize the entire proceedings surrounding the case at hand, they can play an important role for studying the argumentation and course of events. In addition, they detail the majority opinion of the court and the decision, as well as dissenting or concurring opinions by individual judges.

We collected documents from the HUDOC database that are associated with any application number that was attached to a hearing webcast. Due to different document types, there are multiple documents associated with a hearing. Some manual adjustments to the matching were required, because the official database occasionally links to an incorrect document. We exclusively retrieved English documents, but it is also possible to find the link to a French document, should it be available.
During the process of acquiring the matching judgment files, we noticed some of the applications were deemed inadmissible or were withdrawn by the applicant after the hearing. Because having a hearing transcript without a conclusion in form of a judgment appeared insufficient, we excluded all hearings without a judgment file. The list of excluded hearings can be seen in Appendix~\ref{tab:hearings_excluded} along with the reason for exclusion.

All judgment files are provided with their respective case-id and application number, the latter of which serves as a link to the hearing and webcast-id. A webcast only has a single id, while there can be more than one application number connected to a hearing or a case. It should also be noted that there is overlap\footnote{We found an overlap based on the item-id for 190 out of 469 documents} of these documents to the corpus by \citet{chalkidisNeuralLegalJudgment2019}, but unfortunately there is no overlap to the documents annotated by \citet{habernalMiningLegalArguments2023}. This is primarily because LaCour! only contains documents with an importance score of 1, which are key cases and only from a limited date range. The case-ids and application numbers can also serve as a key to identify and collect more material, because it is the same identification as used by the court.

\paragraph{Additional quality control and limitations of current transcripts in LaCour!}
Our transcriptions are not meant to reflect verbatim what is said in the audio, but to aid the understanding of the hearing in text form. We therefore instructed our annotators to not correct the text to reflect stuttering, sentences that have been slightly corrected during speaking, filler words or incorrect grammar. But because Whisper relies on an internal language model, the generated sentences are sometimes grammatically correct, even though the actual spoken sentence is not. This is especially true for the parts we did not manually correct, which concerns everything spoken by the roles ``Applicant", ``Government" and ``Third Party". 
Manually corrected and modified parts are sections spoken by ``Judge" and ``Judge President", as well as all language tags. The remainder of the text has been left untouched and is a direct result of Whisper. Several format and completeness checks were additionally run on the files to ascertain that no tags are incomplete or missing and the judges' names match the ones listed as attendants. On top of that, we automatically checked all transcripts for unusual characters and repeated sentences resulting from looping behavior in the Whisper generative language model.

\paragraph{Dataset accessibility for NLP and explorative studies}
Easy dissemination and readability has a high priority for us, so we provide LaCour! in different file formats.  For comfortable reading, we refer to the concatenated version\footnote{\lacourgithubcorpus} or our online preview page.\footnote{\lacourpreview} The sentence aligned version is available in xml or text format on huggingface.\footnote{\lacourhuggingface}
At this point we would like to emphasize that the transcripts are unofficial interpretations of the hearing webcasts. The webcasts in conjunction with the provided documents are the only official presentation of the proceedings. 

\subsection{LaCour! statistics}
\begin{figure}
	\centering
	\begin{minipage}[t]{0.48\linewidth}
	     \includegraphics[width=1\textwidth]{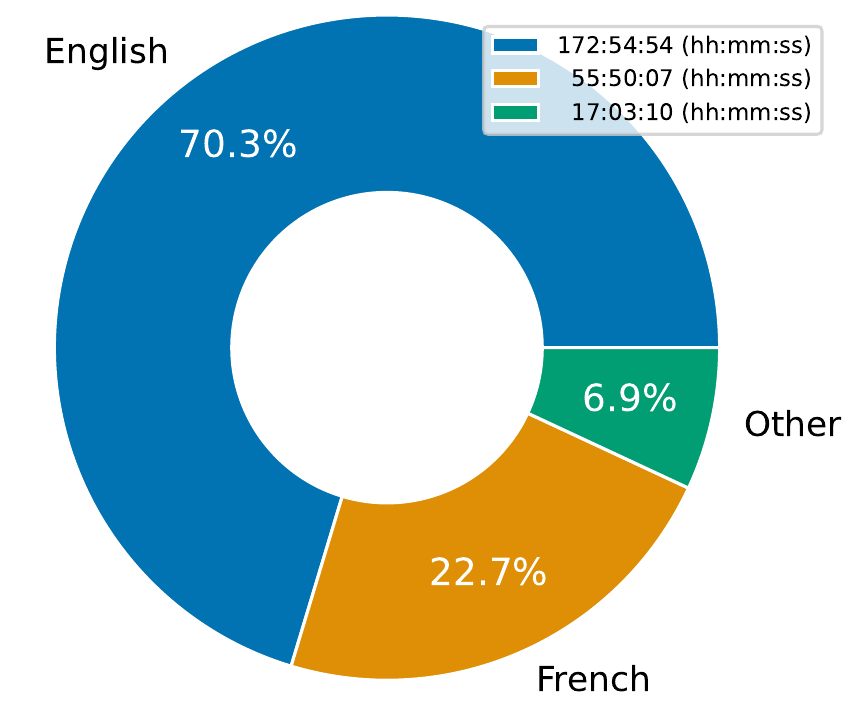}
	      \subcaption{Share of main languages based on time}
	      \label{fig:lang-big}
	\end{minipage}
	\begin{minipage}[t]{0.48\linewidth}
	    \includegraphics[width=1\textwidth]{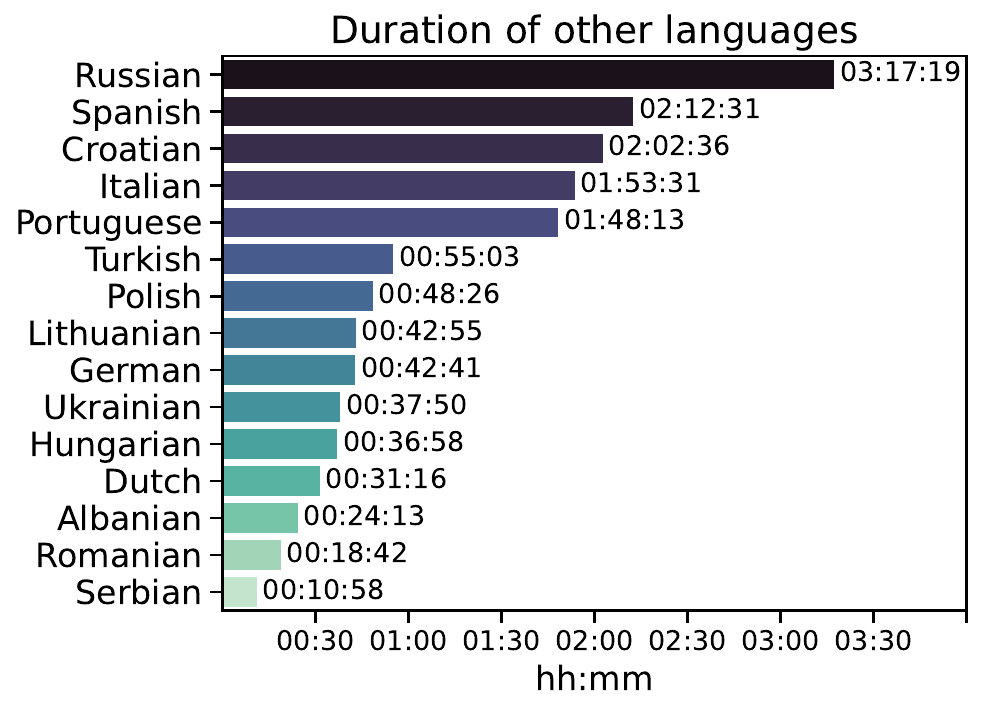}
	    \subcaption{Duration of other languages in the share ``Other"}
	    \label{fig:lang-small}
        \end{minipage} \\	
	\begin{minipage}[t]{0.48\linewidth}
	    \includegraphics[width=1\textwidth]{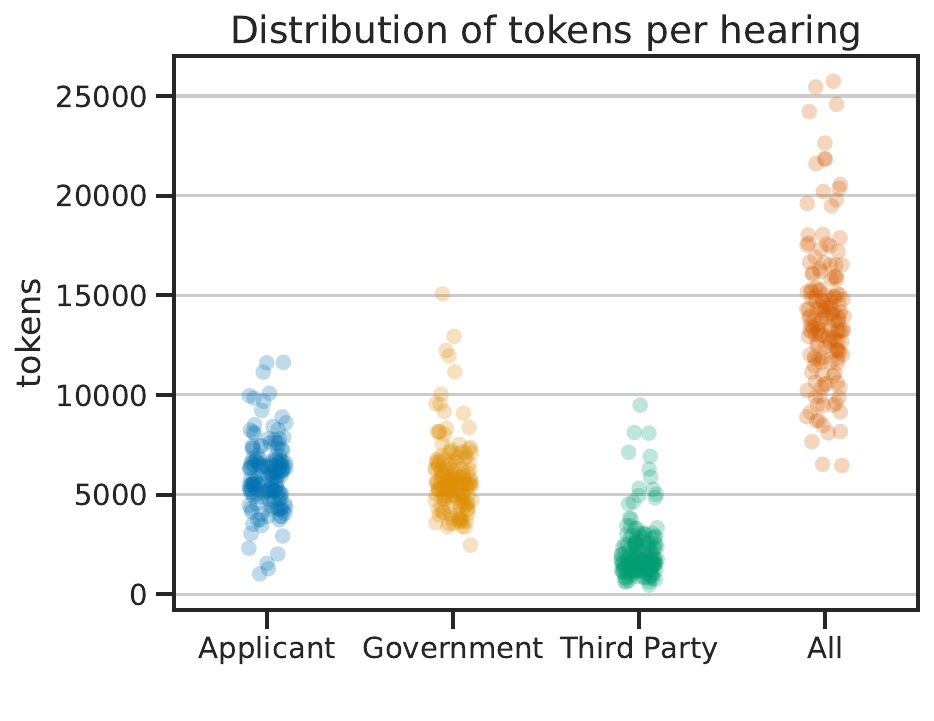}
	    \subcaption{Distribution of the amount of tokens of each hearing grouped by role label}
	\label{fig:distrib-tokens}
         \end{minipage}
         \begin{minipage}[t]{0.48\linewidth}
	    \includegraphics[width=1\textwidth]{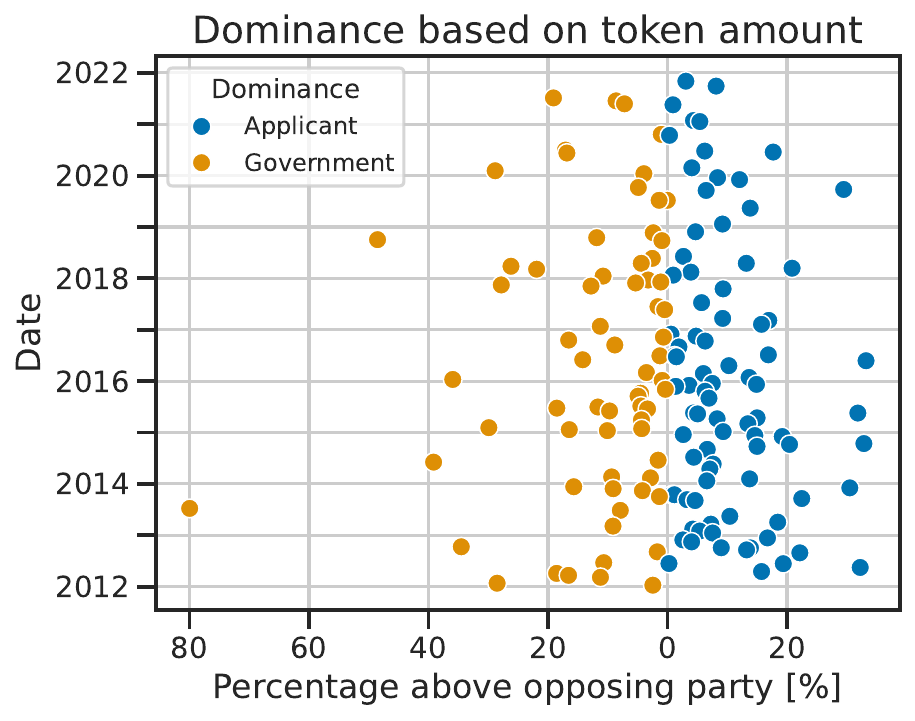}
	    \subcaption{Dominance one party has over the other (absolute) based on the share of tokens}
	\label{fig:dominance}
         \end{minipage}
\caption{Distribution of the languages and roles in the corpus based on time (a,b) and amount of tokens (c,d)}\label{fig:stats}
\end{figure}

The final corpus of hearing transcripts contains 154 hearings ranging from 2012 to 2021 transcribed in their original version and linked to the respective final judgment documents. The total duration of the video content is 267 hours, 20 minutes and 33 seconds with a file size of 153 GB. All 154 transcripts together contain 2,161,833 word tokens. Based on duration of the speeches, a large portion of the transcripts are in English, with French as the second most frequent language. The distributions are illustrated in Figure~\ref{fig:lang-big} with the duration of other languages detailed in Figure~\ref{fig:lang-small}. 

On average, each hearing contains 14,038 word tokens and has an average sentence count of 557. With a standard deviation of 3,519 tokens, the majority of the transcripts are roughly of equal length. 
More descriptive statistics can also be found in Appendix~\ref{sec:appendix-statistics}.
Looking at the amount of tokens per hearing categorized by party outlined in Figure~\ref{fig:distrib-tokens}, one can see that there are several hearings with a larger than average amount of tokens and very few hearings with a comparably smaller amount of tokens. If a third party is involved in the hearing, their part will usually be short. For applicant and government, the amount is also roughly equal. A closer look at the balance between applicant and government is also illustrated in Figure~\ref{fig:dominance}, which shows the absolute difference between the percentage of tokens spoken by the two opposing parties based on the length of the entire hearing. Apart from a few outliers, 85\% of all hearings remain within a maximum of 20\% difference between the two opposing parties. This means that generally, hearings have a balanced amount of time allocated for each party. The outliers can mostly be traced to the presence of more than one party on either side, which means that either there were multiple applicants or multiple respondents involved with fairly allocated speech time. 

\begin{figure}
	\centering
	\begin{minipage}[t]{0.48\linewidth}
	     \includegraphics[width=0.95\textwidth]{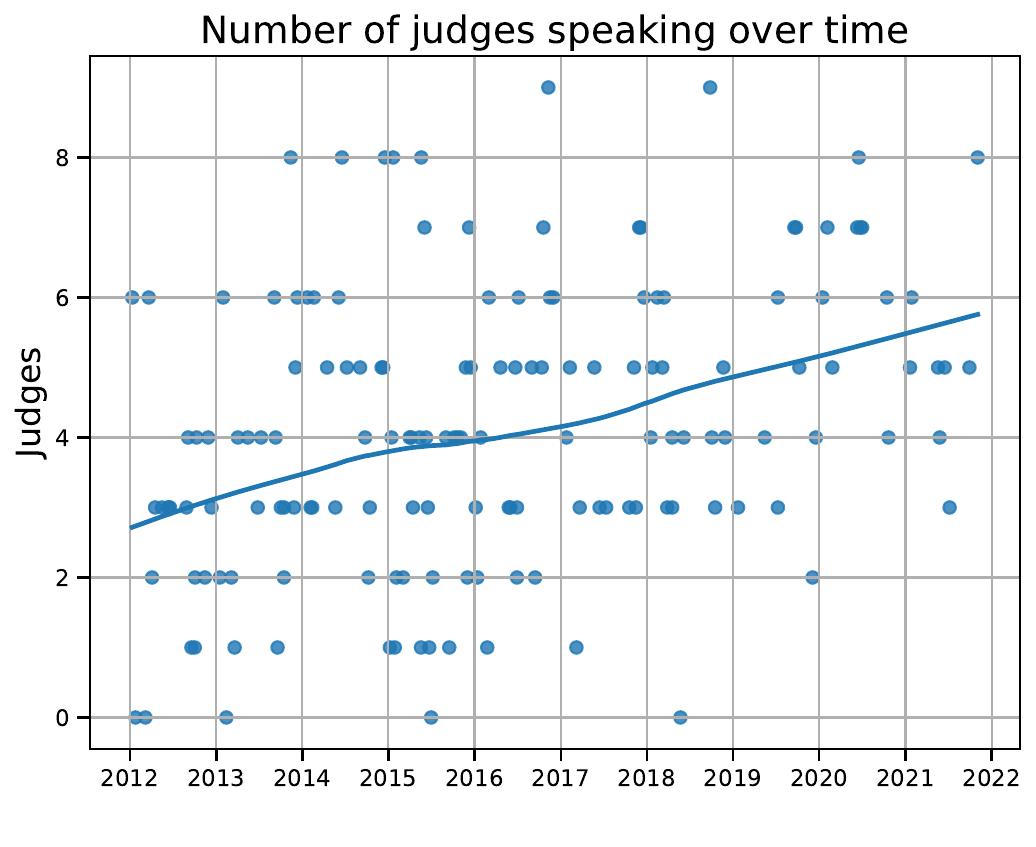}
	      \subcaption{Number of judges asking questions during the interactive part of the hearing over time}
	      \label{fig:judge_questions_amount}
	\end{minipage}
	\begin{minipage}[t]{0.48\linewidth}
	    \includegraphics[width=1\textwidth]{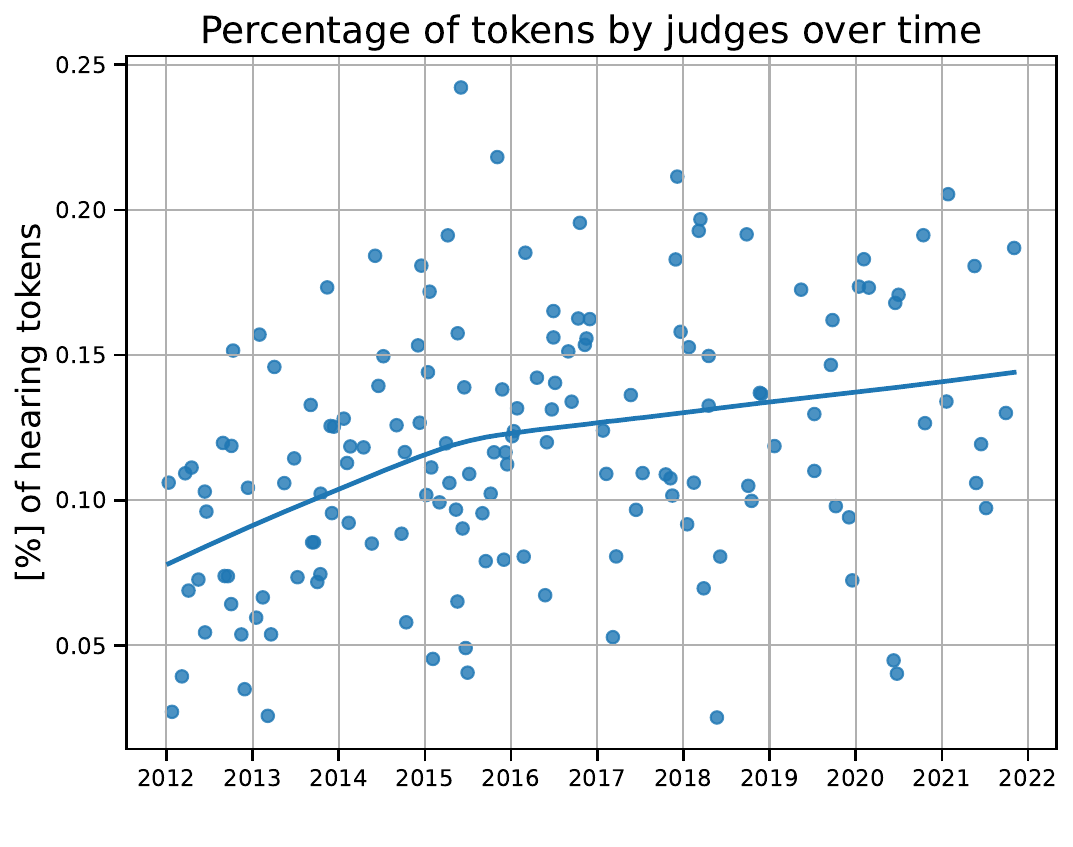}
	    \subcaption{Percentage of tokens spoken by judges in the interactive part of the hearing over time}
	    \label{fig:judge_questions_tokens}
        \end{minipage} \\
\caption{Statistical overview of the hearing structure concerning the interaction and involvement of judge questions.}\label{fig:questions_trend}
\end{figure}

With the labels that we introduced to separate speakers, we can derive a hearing structure and separate parts of the hearing that were spoken before and after the break. The break is usually indicated by the announcer, although if this part is cut out of the video, it is still indicated by the judge president calling on the other judges for questions. With the labels we can also take a look at how the hearing structure has evolved over time. With the hearing structure, we can see that the number of questions asked by judges during hearings has increased over time. The trend can be observed in Figure~\ref{fig:judge_questions_amount}. This is additionally supported by the percentage of tokens by judges in the question session of the hearing over time, shown in Figure~\ref{fig:judge_questions_tokens}, which is also increasing.

\section{Experiments}
LaCour! opens up many new possible points of investigation for ECHR proceedings.
Our new corpus provides us with a window into the argumentation that was used before the judgment was formulated. Apart from the submissions by the parties, we also have insight into the interactions between judges and the parties. This motivates us to investigate the impact of a hearing on the final outcome.

\subsection{Motivation and background}

Besides the presentation of arguments by each party, which are prepared long before the hearing, there is a question section in the hearings. Judges may ask questions to the parties after the submissions and the parties prepare and present answers to these questions. The questions usually ask for some clarifications on specific issues. We have to assume that by the time the hearing takes place, the judges are familiar with the material and circumstances. The question session gives them the opportunity to ask for facts or reasoning by the parties and subliminally point out potential open issues or uncertainties. As indicated in the statistical analysis, the amount of questions has increased in the hearings over time, suggesting that it has become more important for the judges to interact with the parties.

\citet{mathenDecodingCourtLegal2024} thoroughly examined the importance and role of judicial opinions in the Canadian Supreme Court. They found that opinions play an important role and shape the value of a case outcome as a reference for future cases. Opinions are also used to mention and discuss key issues and points of ambiguity and disagreement on the interpretation of the law.

The importance of questions in legal proceedings has also been studied by~\citet{johnsonInquiringMindsWant2009} in the context of the U.S. Supreme Court. This work also brings up the empirical question whether it is possible to detect a judge's stance by their questions. The creation of linked corpora like Super-SCOTUS~\citep{fang-etal-2023-super} also enabled the empirical study of court hearings. \citet{fang-etal-2023-votes} observe that it is possible to detect linguistic clues to a judge's stance in their questions during the oral arguments. They achieve an accuracy of 85\% on the classification of partisanship by training a BERT~\citep{devlin-etal-2018-bert} model using only questions and text spoken by individual justices and the voting behavior.

This brings up the question we want to answer with our experiments. ``Is a judge's opinion on a case already made up by the time of the hearing and does the question give away hints to this opinion?" and ``Is there a connection between a judge having a question during the hearings and a judge having a dissenting or concurring opinion in the subsequent judgment?" There is no partisanship in the ECHR, yet it is possible that we can detect clues that hint at disagreement with the majority.

\subsection{Experimental setup}
We approach the experiments in a similar way to~\citep{fang-etal-2023-votes}, by using only the questions by the judges as input and testing if the question alone gives away hints towards a judge's position. For labels, we consider different configurations, the easiest being the type of opinion that is formulated by the same judge later. We extract all questions from the hearings, which we can easily select by considering all text tagged with the role label ``Judge". To connect the questions with the potential opinions attached to the judgment, we first link the hearing to the correct judgment file using the webcast-id and the type of the hearing (Chamber or Grand Chamber). From the connected judgment file, we extract all opinions attached after the actual judgment declaration. Using the opinion title, which contains the names of the judges who have written the opinion, we can extract the names as labels with regular expressions as well as the type of opinion (``dissenting", ``concurring", ``partly", ``opinion"). We limit the subset to only consider questions in English. After removing the name labels for anonymity, we end up with a sub dataset\footnote{The dataset is available on \lacourhuggingfaceqando} containing 475 individual questions linked to a matching opinion (or none).

We train various models from the BERT~\citep{devlin-etal-2018-bert} family to classify the questions by themselves with the matching label. We chose the basic BERT model to compare it to the results shown in ~\citep{fang-etal-2023-votes}, Legal-BERT~\citep{chalkidisLEGALBERTMuppetsStraight2020}, assuming that it will have a decent performance due to the pretraining on texts from the legal domain, Legal-RoBERTa~\citep{chalkidis-garneau-etal-2023-lexlms} for the same reason and RoBERTa~\citep{liu-etal-2019-roberta}. For the RoBERTa models, we chose the large variant, to see if the model size affects the performance. Additionally, we train a classifier based on the state-of-the-art large language model Llama-3 8b~\citep{llama3modelcard} using 4bit quantization with QLoRA~\citep{dettmers-etal-2024-qlora}. 
The number of instances with the label ``partly" and ``opinion" are severely underrepresented, so we remove them. The remaining labels are ``dissenting", ``concurring" and ``none" with 123, 50 and 282 instances respectively. For the final dataset with 455 instances, we use 70\% for training, 15\% for validation and reserve the remaining 15\% for testing. Due to the imbalanced dataset, we preserve the distribution in the splits and use additional class weights in the training process to avoid overfitting. After the first round of experiments, we reduce the labels further to only classify in a binary manner if there is an opinion or not. 

\subsection{Results}
The training details are shown in Appendix~\ref{appendix:training} and the results can be seen in Table~\ref{tab:results}. 
In the first setting, all models produce similar results, surpassing the random baseline, but not reaching a good stability as indicated by the high standard deviation. Especially the larger models struggle and none of the tested models produce decent results. Looking at the confusion matrix reveals that none of the instances from the class ``concurring" were classified correctly. Surprisingly, Llama-3 8b does not manage to achieve scores that could be considered better than random guessing.
The second setting appears to improve the performance of the models slightly. All models except Llama-3 8b surpass the random baseline. For the smaller models, the domain specific Legal-BERT has an advantage over BERT. But this difference in performance is not observable for the RoBERTa models. The best model in this setting is RoBERTa with a low standard deviation and 65\% F1-score. We will provide an interpretation of these results later in section \ref{sec:analysis.experiments}.

\begin{table}
\centering
\begin{tabular}{llrr}
\toprule
Setting & Model         & F1-score ($\pm$ std)    & Accuracy  ($\pm$ std)  \\
\midrule
\multirow{6}{*}{\shortstack[l]{dissenting, \\ concurring, none}} & Random        & $0.35 \pm 0.03$ & $0.25 \pm 0.01$ \\
&BERT (base)         & $0.44 \pm 0.04$ & $0.35 \pm 0.03$  \\
&Legal-BERT (base)   & $\textbf{0.45} \pm 0.05$ & $\textbf{0.35} \pm 0.09$  \\
&RoBERTa (large)      & $0.41 \pm 0.04$ & $0.33 \pm 0.07$  \\
&Legal-RoBERTa (large) & $0.41 \pm 0.05$ & $0.35 \pm 0.05$  \\
&Llama-3 8b    & $0.32 \pm 0.10$ & $0.24 \pm 0.10$ \\
\midrule
\multirow{6}{*}{\shortstack[l]{dissenting, none}}&Random        & $0.41 \pm 0.02$ & $0.36 \pm 0.03$  \\
&BERT (base)         & $0.56 \pm 0.05$ & $0.48 \pm 0.05$ \\
&Legal-BERT (base)    & $0.62 \pm 0.08$ & $0.55 \pm 0.05$ \\
&RoBERTa (large)      & $\textbf{0.65} \pm 0.01$ & $\textbf{0.58} \pm 0.01$ \\
&Legal-RoBERTa (large) & $0.63 \pm 0.06$ & $0.55 \pm 0.08$ \\
&Llama-3 8b    & $0.51 \pm 0.07$ & $0.52 \pm 0.07$ \\
\bottomrule
\end{tabular}
\caption{Results comparing all models using the labels ``dissenting", ``concurring", ``none" in the first setting and ``dissenting" and ``none" in the second setting. Accuracy refers to the balanced accuracy. Both metrics are averaged over 3 runs.}
\label{tab:results}
\end{table}

\subsection{Question embeddings and topics}
Complementary to the classification experiments, we take a closer look if we can obtain more insights into the differences between questions \emph{with} and \emph{without} a corresponding judge's opinion later on.
To study this, we train a sentence transformer model~\citep{reimers-gurevych-2019-sentence} based on Legal-BERT to embed the question. As training data, we use triplets consisting of a question, a positive example and a negative example. In our case, a positive example is a different question from the same class, a negative example is a different question from a different class. The classes are assigned by the presence and type of opinion of the judge in the same case in which the question was asked: ``none", ``dissenting" and ``concurring". The underlying idea is that questions that appear in the same hearing and are followed up with the same type of opinion should be closer to each other, while questions that appear in the same hearing and resulted in different opinions should be further away from each other. During training, each question in a triplet is moved closer to its positive example and at the same time further away from its negative example. A sentence embedding reflects this by having a higher cosine similarity with similar instances. We avoid duplicates by removing inverse pairs of questions and positive examples. This contrastive sampling generates a dataset of 792 triplets which we used to train a sentence embedding model. The samples have a distribution similar to the dataset used for classification with 442, 250 and 100 triplets for the labels ``none", ``dissenting" and ``concurring". The resulting fine-tuned embedding model can classify triplets with a surprisingly high cosine accuracy of 97\%. 

\begin{figure}
	\centering
    \includegraphics[width=.9\textwidth]{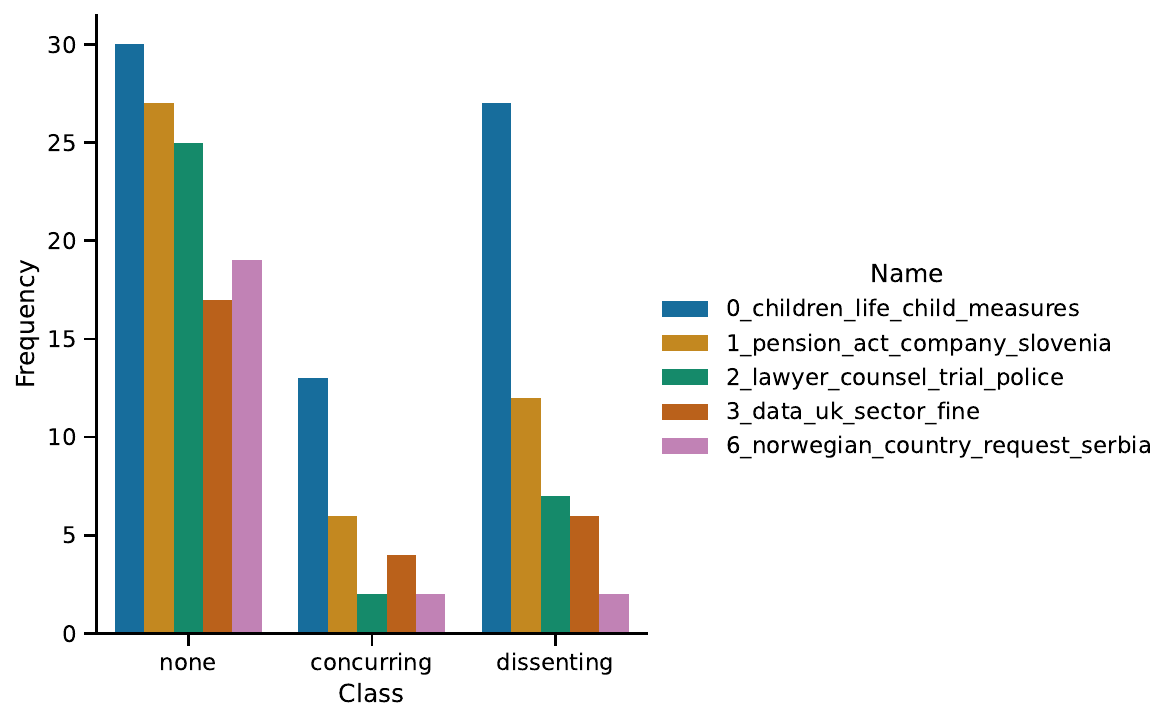}\hfill
	\caption{Selection of the 5 most frequent topics grouped by the opinion type}
	\label{fig:topics_by_class}
\end{figure}

We take a closer look by running a BERT-Topic~\citep{grootendorst2022bertopic} model with these embeddings. To get some meaningful topics, we remove stops words and cluster the embeddings using UMAP~\citep{mcinnes2018umap-software}.
\begin{figure}
	\centering
    \includegraphics[width=1\textwidth]{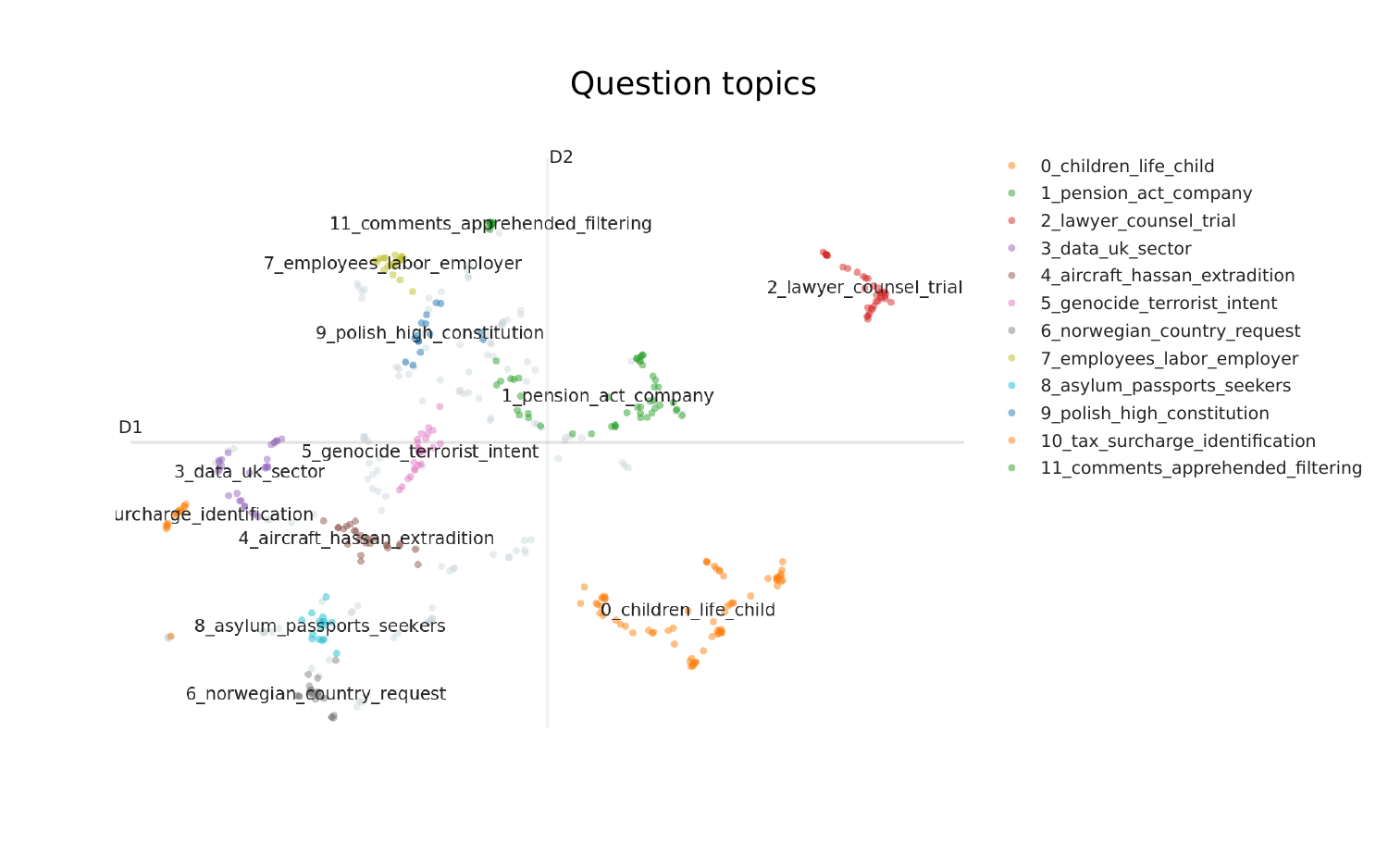}\hfill
	\caption{Clustered topics of all questions. Names are determined by the most frequent words in the cluster. The embeddings are represented on the 2 dimensions that they are reduced on for clustering.}
	\label{fig:topics}
\end{figure}
The topics for all questions are displayed in Figure~\ref{fig:topics}. When taking a look at the topics by class in Figure~\ref{fig:topics_by_class}, we can observe that questions that have a dissenting opinion are more represented in some topics than they are for questions without an opinion. One such example would be the topic ``children\_life\_child", which appears almost as frequently in questions tagged with dissenting opinions as in questions tagged with no opinions, despite the higher amount of questions with no opinion. Although it would require an expert to carefully examine the content of the questions to confirm this observation, this could hint at potential topics which are more controversial than others. Instead of linguistic clues in the way a judge asks their question, the indicating factor could also be linked to specific topics. We believe that this area could be further investigated.

\subsection{Analysis}
\label{sec:analysis.experiments}
Despite high expectations because of the promising results on the similarly formulated task in~\citep{fang-etal-2023-votes}, we could not find such an overwhelming evidence that the questions asked by judges during the hearing can give us foreshadowing to an opinion or voting behavior. Some models managed to surpass the random baseline, yet the training is not stable for all models. Llama-3 8b was even unable to learn anything from the training data. If there is in fact a hint in those questions, it is likely too subtle to be picked up by state-of-the-art language models. Simply distinguishing between ``dissenting" and ``none" instead of the different types of opinions improves the performance of the models. Perhaps the difference between a concurring and a dissenting opinion is too difficult to detect. Alternatively it is possible that the number of examples for questions that resulted in a concurring opinion is too low. We observe that the relatively low performance actually sheds a positive light on the oral hearings in the ECHR, because it suggests that the language used is not as biased as it is in the U.S. Supreme Court. Similar to what~\citet{fang-etal-2023-votes} find, we can also see that a legal domain-specific model does not necessarily work better than a general model. Our best performing model RoBERTa managed to achieve 65\% F1-score on the task of classifying a question that is followed up by a dissenting opinion. These results show that the questions might not be completely free of hints and foreshadowing after all.

Through our experiments using sentence embeddings to represent the question topics, we find that dissenting opinions have certain topics in which they are more represented. This suggests that there are topics which are more controversial in the ECHR and on which the judges disagree more often. Our experiments demonstrated that material which we can extract from the hearings can bear relevancy to the final outcome and we are able to draw a connection between them.

An idea to further investigate this could be the inclusion of audio or video material to detect emotionality. Another possibility would be the inclusion of more transcript material to put the question into context. Yet another approach, which would likely require additional labeled data, could isolate the key issues that the judge's questions address and compare them to the arguments brought forth by both parties on this particular issue. It could be interesting to see which issues are highlighted by judges, how they are addressed by the parties and what the final view of the court is on the issue. A more fine-grained approach like this would dive deeper into the exact argumentation used in hearings. However, this goes beyond the questions that we attempted to answer with our experimental setup, so we leave this for future work. 

\section{Conclusion}
We introduced our new corpus \textbf{LaCour!} which considers oral court proceedings in the ECHR for the first time. This corpus is available to the research community and enables the empirical analysis of ECHR court hearings. In addition to the transcribed and partly manually corrected text from the audio, we provided sentence-level timestamps and manually annotated role and language labels. Our collected data aims to partly close the gap that exists when considering ECHR proceedings as a whole, because it provides insights into the processes that takes place between the submission of an application and the final decision. The dataset characteristics can also enrich existing datasets or enable the formation of new datasets, which we showcased with our subset of questions and opinions. LaCour! will also enable the study of argumentation in hearings, because it contains both the parties' submissions and replies, as well as the questions posed by judges. This difference in modality and style of legal argumentation will also open up the possibility to study interesting novel questions, such as which arguments or views end up in the final judgment.

Although our experiments were limited to using only the questions from the interactive part of the hearing, there are indications that there is possibly a connection between the questions asked in a hearing and opinions on a judgment. We believe that these experiments could be expanded on and our findings warrant further investigation. Apart from the use cases in legal NLP, we also hope that law students or other interested parties will use our data as an educational resource.

\section*{Limitations}
There are some general limitations which originate from the material we worked with. One such limitation is that even though we try to illustrate most of the  process of an application and case, we chose to use the original audio presented at the associated hearing. Although in our opinion, this is the best choice for understanding the speeches by applicants who did not speak English (or French), it should be noted that the judges and attendants of the hearings do not necessarily listen to the original audio. There is no way of knowing which interpretation a judge has heard, it could be either French or English. It is therefore not possible to draw conclusions from the choice of words used in the speeches and the judgments or opinions given by the judges of the Court.
All names, institutions and parties were taken from the official documents provided in the HUDOC database. 
The only sections which are tagged with a name are those spoken by either ``Judge" or ``Judge President". This enables linking questions to opinions in the judgment documents. However, these tags should not be used to automatically profile individual judges.
We would also like to repeat that a large part of the transcript is the uncorrected output of a Whisper model. It is not unlikely that named entities and legal vocabulary are incorrectly transcribed. The quality of the transcriptions has not been evaluated for languages other than English and French.

Concerning the connection between a question and an opinion, there is a factor to be considered, which is the actual decision of violation or non-violation of an article. The composition of the vote is not provided by the court, so we can only assume that a dissenting opinion indicates a vote against the majority. While this could be used as a feature in our study, there is often more than a single vote and multiple articles are concerned in a case. Consequently, the opinion can also be very nuanced and multifaceted which requires a model that can take into account both the fine-grained decision and the content of the opinion.
The topics presented in Figure~\ref{fig:topics_by_class} only represent the questions on a surface level and capture keywords used. Additionally, the exact keywords are not extracted in a fully deterministic manner, so variance is possible.
It should also be noted that under no circumstances do we wish to replicate or emulate a judge. We strongly believe that judgments concerning the lives of people should also be at the hand of an actual person.

\section*{Ethics statement}

This research has been approved by the ethical review board of the Technical University of Darmstadt.

\section*{Acknowledgements}
This work has been supported by the German Research Foundation as part of the ECALP project (HA 8018/2-1), by the hessian.AI ConnectCom grant, and by the Research Center Trustworthy Data Science and Security (\url{https://rc-trust.ai}), one of the Research Alliance centers within the \url{https://uaruhr.de}. We would also like to thank our research assistants who helped with the creation of the LaCour! corpus: Laura Boyette, Yassine Thlija, Marwa Naibkhil, and Leonard Bongard.

\bibliography{bibliography}

\begin{thebibliography}{34}
\providecommand{\natexlab}[1]{#1}
\providecommand{\url}[1]{{#1}}
\providecommand{\urlprefix}{URL }
\providecommand{\doi}[1]{\url{https://doi.org/#1}}
\providecommand{\eprint}[2][]{\url{#2}}
 \bibcommenthead

\bibitem[{AI@Meta(2024)}]{llama3modelcard}
AI@Meta (2024) Llama 3 model card.
  \urlprefix\url{https://github.com/meta-llama/llama3/blob/main/MODEL_CARD.md}

\bibitem[{Aletras et~al(2016)Aletras, Tsarapatsanis, {Preo{\c t}iuc-Pietro},
  and Lampos}]{aletrasPredictingJudicialDecisions2016}
Aletras N, Tsarapatsanis D, {Preo{\c t}iuc-Pietro} D, et~al (2016) {Predicting
  Judicial Decisions of the European Court of Human Rights: A Natural Language
  Processing Perspective}. PeerJ Computer Science 2:e93.
  \doi{10.7717/peerj-cs.93}

\bibitem[{Ashley et~al(2007)Ashley, Pinkwart, Lynch, and
  Aleven}]{ashleyLearningDiagrammingSupreme2007}
Ashley K, Pinkwart N, Lynch C, et~al (2007) {Learning by Diagramming Supreme
  Court Oral Arguments}. In: Proceedings of the 11th International Conference
  on {{Artificial}} Intelligence and Law. Association for Computing Machinery,
  New York, NY, USA, {{ICAIL}} '07, pp 271--275, \doi{10.1145/1276318.1276370}

\bibitem[{Baevski et~al(2020)Baevski, Zhou, Mohamed, and
  Auli}]{baevski-2020-wav2vec2-neurips}
Baevski A, Zhou Y, Mohamed A, et~al (2020) wav2vec 2.0: A framework for
  self-supervised learning of speech representations. In: Larochelle H, Ranzato
  M, Hadsell R, et~al (eds) Advances in Neural Information Processing Systems,
  vol~33. Curran Associates, Inc., pp 12,449--12,460,
  \urlprefix\url{https://proceedings.neurips.cc/paper_files/paper/2020/file/92d1e1eb1cd6f9fba3227870bb6d7f07-Paper.pdf}

\bibitem[{Bergam et~al(2022)Bergam, Allaway, and
  McKeown}]{bergamLegalPoliticalStance2022}
Bergam N, Allaway E, McKeown K (2022) {Legal and Political Stance Detection of
  SCOTUS Language}. In: Proceedings of the {{Natural Legal Language Processing
  Workshop}} 2022. Association for Computational Linguistics, Abu Dhabi, United
  Arab Emirates (Hybrid), pp 265--275, \doi{10.18653/v1/2022.nllp-1.25}

\bibitem[{Bredin et~al(2020)Bredin, Yin, Coria, Gelly, Korshunov, Lavechin,
  Fustes, Titeux, Bouaziz, and Gill}]{Bredin2020}
Bredin H, Yin R, Coria JM, et~al (2020) {Pyannote.Audio: Neural Building Blocks
  for Speaker Diarization}. In: ICASSP 2020 - 2020 IEEE International
  Conference on Acoustics, Speech and Signal Processing (ICASSP), pp
  7124--7128, \doi{10.1109/ICASSP40776.2020.9052974}

\bibitem[{Chalkidis et~al(2019)Chalkidis, Androutsopoulos, and
  Aletras}]{chalkidisNeuralLegalJudgment2019}
Chalkidis I, Androutsopoulos I, Aletras N (2019) {Neural Legal Judgment
  Prediction in English}. In: Proceedings of the 57th {{Annual Meeting}} of the
  {{Association}} for {{Computational Linguistics}}. Association for
  Computational Linguistics, Florence, Italy, pp 4317--4323,
  \doi{10.18653/v1/P19-1424}

\bibitem[{Chalkidis et~al(2020)Chalkidis, Fergadiotis, Malakasiotis, Aletras,
  and Androutsopoulos}]{chalkidisLEGALBERTMuppetsStraight2020}
Chalkidis I, Fergadiotis M, Malakasiotis P, et~al (2020) {LEGAL-BERT: The
  Muppets Straight out of Law School}. In: Findings of the {{Association}} for
  {{Computational Linguistics}}: {{EMNLP}} 2020. Association for Computational
  Linguistics, Online, pp 2898--2904, \doi{10.18653/v1/2020.findings-emnlp.261}

\bibitem[{Chalkidis et~al(2022)Chalkidis, Jana, Hartung, Bommarito,
  Androutsopoulos, Katz, and Aletras}]{chalkidisLexGLUEBenchmarkDataset2022}
Chalkidis I, Jana A, Hartung D, et~al (2022) {LexGLUE: A Benchmark Dataset for
  Legal Language Understanding in English}. In: Proceedings of the 60th
  {{Annual Meeting}} of the {{Association}} for {{Computational Linguistics}}
  ({{Volume}} 1: {{Long Papers}}). Association for Computational Linguistics,
  Dublin, Ireland, pp 4310--4330, \doi{10.18653/v1/2022.acl-long.297}

\bibitem[{Chalkidis et~al(2023)Chalkidis, Garneau, Goanta, Katz, and
  S{\o}gaard}]{chalkidis-garneau-etal-2023-lexlms}
Chalkidis I, Garneau N, Goanta C, et~al (2023) {L}e{XF}iles and {L}egal{LAMA}:
  Facilitating {E}nglish multinational legal language model development. In:
  Rogers A, Boyd-Graber J, Okazaki N (eds) Proceedings of the 61st Annual
  Meeting of the Association for Computational Linguistics (Volume 1: Long
  Papers). Association for Computational Linguistics, Toronto, Canada, pp
  15,513--15,535, \doi{10.18653/v1/2023.acl-long.865},
  \urlprefix\url{https://aclanthology.org/2023.acl-long.865}

\bibitem[{Dettmers et~al(2023)Dettmers, Pagnoni, Holtzman, and
  Zettlemoyer}]{dettmers-etal-2024-qlora}
Dettmers T, Pagnoni A, Holtzman A, et~al (2023) Qlora: Efficient finetuning of
  quantized llms. In: Oh A, Naumann T, Globerson A, et~al (eds) Advances in
  Neural Information Processing Systems, vol~36. Curran Associates, Inc., pp
  10,088--10,115,
  \urlprefix\url{https://proceedings.neurips.cc/paper_files/paper/2023/file/1feb87871436031bdc0f2beaa62a049b-Paper-Conference.pdf}

\bibitem[{Devlin et~al(2018)Devlin, Chang, Lee, and
  Toutanova}]{devlin-etal-2018-bert}
Devlin J, Chang M, Lee K, et~al (2018) {BERT:} pre-training of deep
  bidirectional transformers for language understanding. CoRR abs/1810.04805.
  \urlprefix\url{http://arxiv.org/abs/1810.04805},
  {\href{https://arxiv.org/abs/1810.04805}{{https://arxiv.org/abs/arXiv:1810.04805}}}

\bibitem[{Dickinson(2018)}]{dickinsonComputationalAnalysisOral2018}
Dickinson GM (2018) {A Computational Analysis Of Oral Argument In The Supreme
  Court.} Cornell Journal of Law and Policy 28(3):449

\bibitem[{Fang et~al(2023{\natexlab{a}})Fang, Cohn, Baldwin, and
  Frermann}]{fang-etal-2023-votes}
Fang B, Cohn T, Baldwin T, et~al (2023{\natexlab{a}}) More than votes? voting
  and language based partisanship in the {US} {S}upreme {C}ourt. In: Bouamor H,
  Pino J, Bali K (eds) Findings of the Association for Computational
  Linguistics: EMNLP 2023. Association for Computational Linguistics,
  Singapore, pp 4604--4614, \doi{10.18653/v1/2023.findings-emnlp.306},
  \urlprefix\url{https://aclanthology.org/2023.findings-emnlp.306}

\bibitem[{Fang et~al(2023{\natexlab{b}})Fang, Cohn, Baldwin, and
  Frermann}]{fang-etal-2023-super}
Fang B, Cohn T, Baldwin T, et~al (2023{\natexlab{b}}) Super-{SCOTUS}: A
  multi-sourced dataset for the {S}upreme {C}ourt of the {US}. In:
  Preo{\textcommabelow{t}}iuc-Pietro D, Goanta C, Chalkidis I, et~al (eds)
  Proceedings of the Natural Legal Language Processing Workshop 2023.
  Association for Computational Linguistics, Singapore, pp 202--214,
  \doi{10.18653/v1/2023.nllp-1.20},
  \urlprefix\url{https://aclanthology.org/2023.nllp-1.20}

\bibitem[{Goldman(1998)}]{goldmanPoliticalScienceMultimedia1998}
Goldman J (1998) {Political Science: Multimedia for Research and
  Teaching\textemdash{{The Oyez Oyez Oyez}} and the History and Politics Out
  Loud Projects}. Social Science Computer Review 16(1):30--39.
  \doi{10.1177/089443939801600105}

\bibitem[{Grootendorst(2022)}]{grootendorst2022bertopic}
Grootendorst M (2022) Bertopic: Neural topic modeling with a class-based tf-idf
  procedure. arXiv preprint arXiv:220305794

\bibitem[{Guha et~al(2023)Guha, Nyarko, Ho, R\'{e}, Chilton, K, Chohlas-Wood,
  Peters, Waldon, Rockmore, Zambrano, Talisman, Hoque, Surani, Fagan, Sarfaty,
  Dickinson, Porat, Hegland, Wu, Nudell, Niklaus, Nay, Choi, Tobia, Hagan, Ma,
  Livermore, Rasumov-Rahe, Holzenberger, Kolt, Henderson, Rehaag, Goel, Gao,
  Williams, Gandhi, Zur, Iyer, and Li}]{guha-etal-2023-legalbench}
Guha N, Nyarko J, Ho D, et~al (2023) Legalbench: A collaboratively built
  benchmark for measuring legal reasoning in large language models. In: Oh A,
  Naumann T, Globerson A, et~al (eds) Advances in Neural Information Processing
  Systems, vol~36. Curran Associates, Inc., pp 44,123--44,279,
  \urlprefix\url{https://proceedings.neurips.cc/paper_files/paper/2023/file/89e44582fd28ddfea1ea4dcb0ebbf4b0-Paper-Datasets_and_Benchmarks.pdf}

\bibitem[{Habernal et~al(2023)Habernal, Faber, Recchia, Bretthauer, Gurevych,
  {Spiecker genannt D{\"o}hmann}, and
  Burchard}]{habernalMiningLegalArguments2023}
Habernal I, Faber D, Recchia N, et~al (2023) {Mining Legal Arguments in Court
  Decisions}. Artificial Intelligence and Law \doi{10.1007/s10506-023-09361-y}

\bibitem[{Henderson et~al(2022)Henderson, Krass, Zheng, Guha, Manning,
  Jurafsky, and Ho}]{hendersonPileLawLearning2022}
Henderson P, Krass M, Zheng L, et~al (2022) {Pile of Law: Learning Responsible
  Data Filtering from the Law and a 256GB Open-Source Legal Dataset}. Advances
  in Neural Information Processing Systems 35:29,217--29,234

\bibitem[{Johnson et~al(2009)Johnson, Black, Goldman, and
  Treul}]{johnsonInquiringMindsWant2009}
Johnson TR, Black RC, Goldman J, et~al (2009) {Inquiring Minds Want to Know: Do
  Justices Tip Their Hands with Questions at Oral Argument in the U.S. Supreme
  Court}. Washington University Journal of Law \& Policy 29:241

\bibitem[{Liu et~al(2019)Liu, Ott, Goyal, Du, Joshi, Chen, Levy, Lewis,
  Zettlemoyer, and Stoyanov}]{liu-etal-2019-roberta}
Liu Y, Ott M, Goyal N, et~al (2019) Roberta: {A} robustly optimized {BERT}
  pretraining approach. CoRR abs/1907.11692.
  \urlprefix\url{http://arxiv.org/abs/1907.11692},
  {\href{https://arxiv.org/abs/1907.11692}{{https://arxiv.org/abs/arXiv:1907.11692}}}

\bibitem[{Mathen et~al(2024)Mathen, Alschner, and
  MacDonnell}]{mathenDecodingCourtLegal2024}
Mathen C, Alschner W, MacDonnell V (2024) Decoding the {{Court}}: {{Legal Data
  Insights}} from the {{Supreme Court}} of {{Canada}}, 1st edn. Routledge,
  London, \doi{10.4324/9781003279112}

\bibitem[{McInnes et~al(2018)McInnes, Healy, Saul, and
  Grossberger}]{mcinnes2018umap-software}
McInnes L, Healy J, Saul N, et~al (2018) Umap: Uniform manifold approximation
  and projection. The Journal of Open Source Software 3(29):861

\bibitem[{Medvedeva et~al(2020)Medvedeva, Vols, and
  Wieling}]{medvedevaUsingMachineLearning2020}
Medvedeva M, Vols M, Wieling M (2020) {Using Machine Learning to Predict
  Decisions of the European Court of Human Rights}. Artificial Intelligence and
  Law 28(2):237--266. \doi{10.1007/s10506-019-09255-y}

\bibitem[{Mochales and Ieven(2009)}]{mochalesCreatingArgumentationCorpus2009}
Mochales R, Ieven A (2009) {Creating an Argumentation Corpus: Do Theories Apply
  to Real Arguments? A Case Study on the Legal Argumentation of the ECHR}. In:
  Proceedings of the 12th {{International Conference}} on {{Artificial
  Intelligence}} and {{Law}}. Association for Computing Machinery, New York,
  NY, USA, {{ICAIL}} '09, pp 21--30, \doi{10.1145/1568234.1568238}

\bibitem[{Mochales and Moens(2011)}]{mochalesArgumentationMining2011a}
Mochales R, Moens MF (2011) {Argumentation Mining}. Artificial Intelligence and
  Law 19(1):1--22. \doi{10.1007/s10506-010-9104-x}

\bibitem[{{Mochales-Palau} and
  Moens(2007)}]{mochales-palauStudySentenceRelations2007}
{Mochales-Palau} R, Moens MF (2007) {Study on Sentence Relations in the
  Automatic Detection of Argumentation in Legal Cases}. In: Proceedings of the
  2007 Conference on {{Legal Knowledge}} and {{Information Systems}}: {{JURIX}}
  2007: {{The Twentieth Annual Conference}}. IOS Press, NLD, pp 89--98

\bibitem[{Poudyal et~al(2020)Poudyal, Savelka, Ieven, Moens, Goncalves, and
  Quaresma}]{poudyalECHRLegalCorpus2020}
Poudyal P, Savelka J, Ieven A, et~al (2020) {ECHR: Legal Corpus for Argument
  Mining}. In: Proceedings of the 7th {{Workshop}} on {{Argument Mining}}.
  Association for Computational Linguistics, Online, pp 67--75

\bibitem[{Radford et~al(2023)Radford, Kim, Xu, Brockman, Mcleavey, and
  Sutskever}]{radford2022whisper}
Radford A, Kim JW, Xu T, et~al (2023) {Robust Speech Recognition via
  Large-Scale Weak Supervision}. In: Krause A, Brunskill E, Cho K, et~al (eds)
  Proceedings of the 40th International Conference on Machine Learning,
  Proceedings of Machine Learning Research, vol 202. PMLR, pp 28,492--28,518,
  \urlprefix\url{https://proceedings.mlr.press/v202/radford23a.html}

\bibitem[{Reimers and Gurevych(2019)}]{reimers-gurevych-2019-sentence}
Reimers N, Gurevych I (2019) {Sentence-BERT: Sentence Embeddings using Siamese
  BERT-Networks}. In: Inui K, Jiang J, Ng V, et~al (eds) Proceedings of the
  2019 Conference on Empirical Methods in Natural Language Processing and the
  9th International Joint Conference on Natural Language Processing
  (EMNLP-IJCNLP). Association for Computational Linguistics, Hong Kong, China,
  pp 3982--3992, \doi{10.18653/v1/D19-1410},
  \urlprefix\url{https://aclanthology.org/D19-1410}

\bibitem[{Spaeth et~al(2022)Spaeth, Epstein, Martin, Segal, Ruger, and
  Benesh}]{Spaeth2022}
Spaeth HJ, Epstein L, Martin AD, et~al (2022) {2022 Supreme Court Database,
  Version 2022 Release 01}. \urlprefix\url{http://Supremecourtdatabase.org}

\bibitem[{Villata et~al(2022)Villata, Araszkiewicz, Ashley, {Bench-Capon},
  Branting, Conrad, and Wyner}]{villataThirtyYearsArtificial2022a}
Villata S, Araszkiewicz M, Ashley K, et~al (2022) {Thirty Years of Artificial
  Intelligence and Law: The Third Decade}. Artificial Intelligence and Law
  30(4):561--591. \doi{10.1007/s10506-022-09327-6}

\bibitem[{Voeten(2008)}]{voetenImpartialityInternationalJudges2008}
Voeten E (2008) The impartiality of international judges: Evidence from the
  {European Court of Human Rights}. American Political Science Review
  102(4):417--433. \doi{10.1017/S0003055408080398}

\end{thebibliography}

\newpage
\appendix

\section{Hearing statistics}
\label{sec:appendix-statistics}
The hearings in Table~\ref{tab:hearings_excluded} were eventually excluded from the corpus:
\begin{table*}[!htbp]
	\centering
	\begin{tabularx}{\textwidth}{l>{\raggedright\arraybackslash}Xl>{\raggedright\arraybackslash}X}
		\toprule
		WebcastID & Title & Date & Reason \\ \midrule
		233009\_07112012 & Sindicatul `Pastorul cel Bun' v. Romania (no. 2330/09) & 07.11.2012 & contains incorrect voice overlay in audio \\
		4221907\_12112014 & Gherghina v. Romania (no. 42219/07) & 12.11.2014 & declared inadmissible (non-exhaustion of domestic remedies) \\
		7153714\_11012017 & Harkins v. the United Kingdom (no. 71537/14) & 11.01.2017 & Interim measure were lifted, application was deemed inadmissible \\
		5842813\_22112017 & Berlusconi v. Italy (no. 58428/13) & 22.11.2017 & case withdrawn by applicant \\
		6312915\_13112018 & Elçi v. Turkey (no. 63129/15), Ahmet Tunç and Others v. Turkey (no. 4133/16) and Tunç and Yerbasan v. Turkey (no. 31542/16) & 13.11.2018 & All applications were deemed inadmissible \\
		359918\_24042019 &	M.N. and Others v. Belgium (no. 3599/18) & 24.04.2019 & application deemed inadmissible\\
		5415516\_12062019 & Slovenia v. Croatia (no. 54155/16) & 12.06.2019 &  no jurisdiction to take cognisance of the application \\
		2095814\_11092019 & Ukraine v. Russia (re Crimea) (no. 20958/14) & 11.09.2019 & decision partly admissible, but another hearing is planned for 2023 \\
		\bottomrule
	\end{tabularx}
	\caption{Hearings which were excluded from the corpus and the reason for the exclusion}
	\label{tab:hearings_excluded}
\end{table*}

More detailed statistics on LaCour! can be seen in Table~\ref{tab:descriptive-stats}.
\begin{table*}[!htbp]
	\centering
	\begin{tabular}{lr}
	\toprule
    number of transcripts & 154 \\
    date range & 01.01.2012 - 31.12.2021\\
    \midrule
	tokens (nltk) & 2,161,833 \\
	range of tokens per transcript & 6458 - 25742 \\
	average tokens per transcript ($\pm \sigma$) & 14,038 $\pm$ 3519\\
	median tokens per transcript & 15,163 \\
	quartile of tokens [25\%, 50\%, 75\%]	& [12020, 13784, 15295] \\
	vocabulary size & 77,156 \\
	average sentences per transcript ($\pm \sigma$) & 557 $\pm$ 155.90\\
	quartile of sentences [25\%, 50\%, 75\%]	& [454, 550, 625] \\
	average tokens per sentence ($\pm \sigma$) & 25.2 $\pm$ 19.50\\
	average conversation turns per transcript ($\pm \sigma$) & 26 $\pm$ 7.68\\
	quantile (0.85) token \% Applicant \textgreater~Government  & 17.98 \\
	quantile (0.85) token \% Government \textless~Applicant  & 21.35 \\
	\bottomrule
	\end{tabular}
\caption{Additional descriptive statistics for the LaCour! corpus}
\label{tab:descriptive-stats}
\end{table*}
\newpage
\section{Experimental setup details}
\label{appendix:training}

The following training setups were considered for our experiments. The full dataset of questions and opinions is also available on huggingface\footnote{\lacourhuggingfaceqando} and contains a partition with 475 pairs of questions and opinions (or lack thereof) with a label for the opinion type for classification. The second partition consists of 792 non-duplicate triplets of a question, a positive example and a negative example used for training a sentence-transformer model with MultipleNegativesRankingLoss.\footnote{\url{https://sbert.net/docs/package_reference/sentence_transformer/losses.html}}

For all classification models, except Llama-3, we train the base models on the training data for 20 epochs with a batch size of 8 and an adaptive learning rate of 2e-6.
The classification model using Llama-3 8b was done with 4bit quantization and a LoRA configuration to reduce the computational requirements for training a Llama model. Training for 3 epochs already lead to overfitting, so we reduced the training to only run for 1 epoch. All configurations use class weights to account for the imbalanced training data. We run all models on 3 different random seeds (75, 953, 5113), average across them and monitor F1-score, balanced accuracy as well as the confusion matrix. 

For the sentence-transformer model we use Legal-BERT as a base model and train for 150 epochs with a batch size of 32. The chosen seed is 5113 and the dataset is split to reflect the class distribution into 70\% for training, 15\% for validation and the remaining 15\% for testing.

We make the code for all experiments available on github.\footnote{\lacourqando}

\end{document}